\documentclass[10pt,conference]{IEEEtran}
\IEEEoverridecommandlockouts

\usepackage[T1]{fontenc}
\usepackage{cite}
\usepackage{url}
\usepackage{amsmath,amssymb,amsfonts}
\usepackage{algorithmic}
\usepackage{graphicx}
\usepackage{textcomp}
\usepackage{xcolor}
\usepackage{fvextra}
\usepackage{subcaption}
\usepackage{booktabs}
\usepackage{pgfplots}
\pgfplotsset{compat=1.18}
\usepackage{array}
\usepackage{xcolor}
\usepackage{multirow}
\usepackage{listings}
\usepackage[table]{xcolor}
\usepackage{tikz}
\usetikzlibrary{positioning,arrows.meta,shapes.geometric}
\usepackage[most]{tcolorbox}
\usepackage{enumitem}
\setlist[enumerate]{wide=0pt, labelindent=0pt, itemsep=2pt, topsep=2pt}
\setlist[itemize]{wide=0pt, labelindent=0pt, itemsep=2pt, topsep=2pt}

\newcommand{\boxmargin}{2mm}
\newtcolorbox{rqboxenv}{
    colback=yellow!10!white,
    colframe=gray!50,
    arc=0pt, outer arc=5pt,
    boxsep=0pt,
    leftrule=3pt, bottomrule=0pt, toprule=0pt, rightrule=0pt,
    left=\boxmargin, right=\boxmargin, top=1mm, bottom=1mm,
    before skip=3pt, after skip=3pt
}
\newcommand{\rqbox}[1]{\begin{rqboxenv}#1\end{rqboxenv}}

\definecolor{diffaddbg}{RGB}{235,255,238}
\definecolor{diffrmbg}{RGB}{255,235,235}
\definecolor{diffaddbar}{RGB}{40,167,69}
\definecolor{diffrmbar}{RGB}{203,36,49}
\definecolor{LightCyan}{rgb}{0.94,1,1}
\definecolor{barblue}{RGB}{68,114,196}
\newcommand{\hlbest}[1]{\cellcolor{green!30}\textbf{#1}}
\newcommand{\hlsecond}[1]{\cellcolor{green!15}\underline{#1}}

\newcommand{\diffrm}[2][0]{
  \ifblank{#2}{}{
    \begingroup
      \setlength{\fboxsep}{1.2pt}
      \noindent\colorbox{diffrmbg}{%
        \parbox{\dimexpr\linewidth-2\fboxsep\relax}{%
          \setlength{\baselineskip}{8pt}%
          \textcolor{diffrmbar}{\ttfamily\scriptsize-}\hspace{0.4em}\hspace*{#1 em}{\ttfamily\scriptsize\detokenize{#2}}%
        }%
      }%
      \par\nointerlineskip
    \endgroup
  }
}

\newcommand{\diffadd}[2][0]{
  \ifblank{#2}{}{
    \begingroup
      \setlength{\fboxsep}{1.2pt}
      \noindent\colorbox{diffaddbg}{%
        \parbox{\dimexpr\linewidth-2\fboxsep\relax}{%
          \setlength{\baselineskip}{8pt}%
          \textcolor{diffaddbar}{\ttfamily\scriptsize+}\hspace{0.4em}\hspace*{#1 em}{\ttfamily\scriptsize\detokenize{#2}}%
        }%
      }%
      \par\nointerlineskip
    \endgroup
  }
}

\usepackage[hidelinks]{hyperref}
\begin{document}
\flushbottom

\title{Beyond Pixel Diffs: Benchmarking Image Change Captioning for Web UI Visual Regression Testing}

\author{\IEEEauthorblockN{Licheng Zhang, Bach Le, Pengtao Zhao, Naveed Akhtar}
\IEEEauthorblockA{\textit{School of Computing and Information Systems}, \textit{University of Melbourne}, Melbourne, Australia \\
\{licheng.zhang@student.unimelb.edu.au, bach.le@unimelb.edu.au,\\ pengtaozhao@unimelb.edu.au, naveed.akhtar1@unimelb.edu.au\}}
}

\maketitle

\begin{abstract}
Visual regression testing (VRT) is a standard quality assurance step in modern software release pipelines. On every change, it re-renders user interface (UI) screenshots, compares each one against an approved baseline image, and routes any detected difference to a human reviewer who decides whether it is an intended update or an unintended regression. A widely used approach, especially in open-source and continuous-integration pipelines, is pixel-level comparison, which is semantically blind and treats rendering noise and genuine defects identically, producing large volumes of false positives that force developers and testers to spend substantial time and effort manually reviewing flagged differences at every release cycle. Industry tools apply machine learning to VRT, but lack public evaluation. More critically, no dataset or benchmark exists to support natural language descriptions of UI changes, a capability that tells testers what changed in words instead of leaving them to interpret a binary flag or a highlighted region. To address the gap, we propose a new task, Web UI Image Change Captioning (WUICC), which sits at the intersection of VRT and image difference captioning (IDC), and release WUICC-bench, its first dataset and benchmark for the task. We build the dataset with a controlled large language model (LLM)-driven mutation pipeline. Each sample applies a single atomic change drawn from a 37-rule taxonomy of Web UI changes, spanning nine meaningful and three non-meaningful categories grounded in VRT practice, and is paired with a natural language caption, yielding 9,906 human-verified samples. We evaluate eleven representative IDC methods, together with two zero-shot general-purpose LLMs. We find that: (1) these methods tend to struggle in the Web UI domain due to its layout diversity, dense text, and fine-grained changes, and (2) yet the trained methods already suppress non-meaningful visual noise far more selectively than the pixel-level comparison VRT relies on, providing a solid foundation for future domain-specific research.
\end{abstract}

\section{Introduction}
\begin{figure*}[t]
\centering
\vspace{-3mm}
\begin{minipage}[t]{0.49\linewidth}
\vspace*{0pt}
\begin{minipage}[t][2.8cm][t]{\linewidth}
  \setlength{\parskip}{0pt}
  \begin{center}
  \begin{tikzpicture}[baseline=(current bounding box.south)]
    \path (0,0) rectangle (3.8,0.86);
    \fill[black!5]
      (1.09,0.09) rectangle (2.71,0.77);
    \fill[rounded corners=0.28cm, white]
      (1.15,0.15) rectangle (2.65,0.71);
    \node[font=\scriptsize\bfseries, text=black] at (1.9,0.43) {Book Now};
  \end{tikzpicture}
  \hspace{0.5mm}\raisebox{0.35cm}{\scriptsize$\rightarrow$}\hspace{0.5mm}
  \begin{tikzpicture}[baseline=(current bounding box.south)]
    \path (0,0) rectangle (3.8,0.86);
    \fill[black!5]
      (0.24,0.09) rectangle (3.44,0.77);
    \fill[rounded corners=0.28cm, white]
      (0.3,0.15) rectangle (1.8,0.71);
    \node[font=\scriptsize\bfseries, text=black] at (1.05,0.43) {Book Now};
    \fill[rounded corners=0.28cm, white]
      (1.88,0.15) rectangle (3.38,0.71);
    \node[font=\scriptsize\bfseries, text=black] at (2.63,0.43) {Learn More};
  \end{tikzpicture}
  \end{center}
  \vspace{-4pt}
  \diffadd{<button class="mx-auto mt-2 lg:mx-0 hover:underline bg-white text-gray-800 font-bold rounded-full my-2 py-3 px-6 shadow-lg">Learn More</button>}
  \par\vspace{2pt}
  \noindent\parbox{\linewidth}{\scriptsize\setlength{\baselineskip}{7pt}\textbf{Change:} A new button with the text ``Learn More'' was added next to the ``Book Now'' button in the hero section.}
\end{minipage}\par\vspace{2pt}
\end{minipage}
\hfill
\begin{minipage}[t]{0.49\linewidth}
\vspace*{0pt}
\begin{minipage}[t][2.8cm][t]{\linewidth}
  \setlength{\parskip}{0pt}
  \begin{center}
  \begin{tikzpicture}[baseline=(current bounding box.south)]
    \path (0,0) rectangle (3.2,0.35);
    \node[font=\tiny\bfseries, text=black, anchor=base west] at (0.3,0.14) {Courses};
    \node[font=\tiny\bfseries, text=black, anchor=base west] at (1.45,0.14) {Degree Programs};
  \end{tikzpicture}
  \hspace{0.5mm}\raisebox{0.145cm}{\scriptsize$\rightarrow$}\hspace{0.5mm}
  \raisebox{0.02cm}{
  \begin{tikzpicture}[baseline=(current bounding box.south)]
    \path (0,0) rectangle (3.2,0.35);
    \node[font=\tiny\bfseries, text=black, anchor=base west] at (0.3,0.14) {Courses};
  \end{tikzpicture}}
  \end{center}
  \vspace{-5pt}
  \diffrm{<a href="#responsive-header" class="block mt-4 lg:inline-block lg:mt-0 text-teal-200 hover:text-white mr-4">}
  \diffrm[1]{Degree Programs}
  \diffrm{</a>}
  \par\vspace{2pt}
  \noindent\parbox{\linewidth}{\scriptsize\setlength{\baselineskip}{7pt}\textbf{Change:} The navigation item with the text ``Degree Programs'' in the header was removed.}
\end{minipage}\par\vspace{2pt}
\end{minipage}
\vspace{-18pt}
\caption{Two representative examples of meaningful Web UI changes in WUICC-bench, shown as schematic illustrations of the changed region for compactness, with only the modified code lines included.}
\vspace{-18pt}
\label{fig:samples}
\end{figure*}
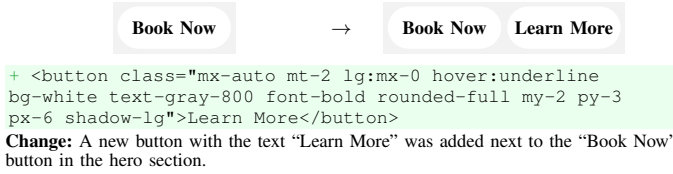

Visual regression testing (VRT) is a standard quality assurance practice in modern software engineering (SE)~\cite{alegroth2017long,mahajan2015detection}. Before each release, user interface (UI) screenshots are captured and compared against an approved baseline to detect unintended visual changes introduced by code, style, or content updates, and every flagged difference is routed to a human reviewer who decides whether it is an intended update or a regression~\cite{adachi2018reducing}. As release cycles shorten and UIs grow more complex~\cite{chen2015continuous}, VRT has become a critical safeguard against user-facing regressions. Yet, pixel-by-pixel comparison, widely used in open-source and continuous-integration pipelines~\cite{vrt_sqa_ml_2025}, is semantically blind. It flags any deviation, treating sub-pixel rendering jitter, anti-aliasing differences, and dynamic advertisement content identically to genuine structural regressions such as missing components, layout restructuring, or altered navigation~\cite{adachi2018reducing,tanno2020region,aridome2026mixvrt}. Such indiscriminate flagging produces a high false-positive rate, forcing developers and testers to manually inspect large volumes of flagged screenshots at every release and making VRT a significant and recurring source of quality assurance overhead~\cite{adachi2018reducing,adachi2020method}. Industry tools such as Applitools Eyes~\cite{applitools} and Percy~\cite{percy} have begun incorporating comparison based on machine learning to reduce false positives, but they are proprietary and lack public evaluation.
Hence, separation of meaningful regression from visual noise remains broadly underexplored. Recent research has begun applying vision-language models (VLMs) to visual software engineering tasks, such as detecting cross-browser inconsistencies~\cite{xbidetective2026} and repairing visual software issues~\cite{huang2025seeing}, yet none produces a natural language description of UI changes for VRT.

Image difference captioning (IDC) is precisely the missing capability in VRT pipelines. Given a pair of related images, an IDC model localizes what changed, infers the semantic nature of the change, and verbalizes it as a fluent natural language description, producing output that is both interpretable and selective in ways that pixel-level signals are not. The task has been studied extensively in remote sensing~\cite{cai2023interactive,li2024inter,xian2025dynamic,xue2026towards,yang2025change,li2026exploring} and natural image domains~\cite{yao2022image,jhamtani2018learning,tu2023adaptive,tu2023self,black2024vixen,liu2025omnidiff}, where it underpins applications ranging from land-cover monitoring~\cite{liu2025remote} to scene-change description~\cite{jhamtani2018learning}. Yet, despite the breadth of prior work, IDC has never been applied to Web UIs. Unlike natural photographs or satellite imagery, which are captured by cameras and characterized by continuous textures and smooth spatial gradients, Web UI screenshots are programmatically rendered artifacts that are highly structured, text-dense, and composed of axis-aligned components with sharp boundaries and strong layout regularity. Unlike the IDC settings above, where the goal is to describe whatever has changed, the Web UI setting inherits VRT's defining concern of telling genuine regressions apart from noise, so its changes split into two kinds, meaningful and non-meaningful. Meaningful changes are genuine regressions that should be reported to a reviewer, spanning element-level modifications, textual updates, layout restructuring, and thematic redesigns, with representative examples shown in Figure~\ref{fig:samples}. Non-meaningful changes, such as sub-pixel shifts or minor style adjustments, are the visual noise that VRT pipelines must suppress. Whether the modeling paradigm that makes IDC effective in natural image and remote sensing domains carries over to the Web UI setting is an open question, and one that cannot be answered without a domain-specific dataset and benchmark.

We take a direct step toward answering it by introducing Web UI Image Change Captioning (WUICC), a new task that applies IDC to the VRT setting, and release WUICC-bench, the first dataset and benchmark for the task. Answering the question requires, at minimum, controlled Web UI image pairs that reflect realistic change distributions, natural language annotations that capture the semantics of those changes, and a systematic evaluation protocol that enables fair comparison across methods. None of these exist for the Web UI domain, and WUICC-bench is designed to provide all three. We generate the data with a controlled large language model (LLM)-driven mutation pipeline over HTML pages from WebSight~\cite{laurenccon2024unlocking}, a public corpus of LLM-synthesized Web UI pages. Each sample applies exactly one atomic change drawn from a taxonomy of Web UI changes grounded in VRT literature~\cite{adachi2018reducing,tanno2020region,aridome2026mixvrt,aridome2025prototype} and industry tooling practice~\cite{applitools,percy}, and is paired with an LLM-generated caption that human annotators verify. The LLM applies a semantically coherent, taxonomy-controlled change and emits an aligned caption in the same pass, so every mutation comes with ground-truth supervision. Because the LLM is conditioned only on HTML, with rendering delegated to a headless browser, the pipeline scales cost-effectively.

To assess whether existing IDC methods transfer to the Web UI domain, we re-implement eleven representative methods, spanning CNN-based, Transformer-based, and Mamba-based architectures originally developed for natural images or remote sensing, and evaluate them on WUICC-bench with fixed data splits, tokenization, and evaluation metrics so that the methods differ only in their modeling of change. Our contributions are as follows.
\begin{enumerate}
    \item We propose WUICC, a new task connecting VRT to IDC, opening a new research direction at the intersection of SE and computer vision and providing a formal problem definition for an important but previously unaddressed SE challenge. We are explicit that the novelty lies in the task framing, the change taxonomy, and the benchmark construction rather than in a new captioning architecture, and we reuse existing IDC methods unchanged so that the reported gap reflects the domain rather than our own modeling choices.
    \item We construct and publicly release WUICC-bench through a controlled and verifiable LLM-driven mutation pipeline over Web UI pages. Each sample applies exactly one atomic change drawn from a 37-rule taxonomy, is paired with an automatically generated natural language caption, and passes human verification. We report the pipeline's success rate and inter-annotator agreement to characterize its reliability. The resulting dataset spans nine meaningful change categories, including element addition and removal, layout restructuring, reordering, and thematic redesigns, together with three non-meaningful categories representing visual noise. Beyond the released data, the pipeline is itself a reusable methodology for building change captioning datasets in structured-UI domains rather than a one-off construction effort.
    \item We develop a Web UI change taxonomy grounded in VRT literature and industry tooling practice, and we make explicit the design criteria we propose for it, namely coverage of the changes VRT encounters, separability between categories, and one atomic change per sample.
    \item We provide the first empirical benchmark for the WUICC task by re-implementing and evaluating eleven representative IDC methods spanning CNN-based, Transformer-based, and Mamba-based architectures, together with two zero-shot general-purpose VLMs. Our results indicate that both the trained methods and the VLMs tend to struggle in the Web UI domain, generally underperforming relative to IDC scores on natural image and remote sensing benchmarks, yet the trained methods already suppress non-meaningful changes far more selectively than pixel-level comparison, the capability VRT most needs. We offer WUICC-bench as a shared foundation for future domain-specific research.
\end{enumerate}

The remainder of the paper is organized as follows. Section~\ref{sec:background} reviews background on VRT and IDC and characterizes the challenges of the WUICC problem. Section~\ref{sec:dataset} describes WUICC-bench. Section~\ref{sec:experiments} presents the experimental setup, results, and ablation studies. Section~\ref{sec:discussion} presents discussion and threats to validity. Section~\ref{sec:related} situates our work within the broader literature, and Section~\ref{sec:conclusion} concludes the paper.

\section{Background}\label{sec:background}
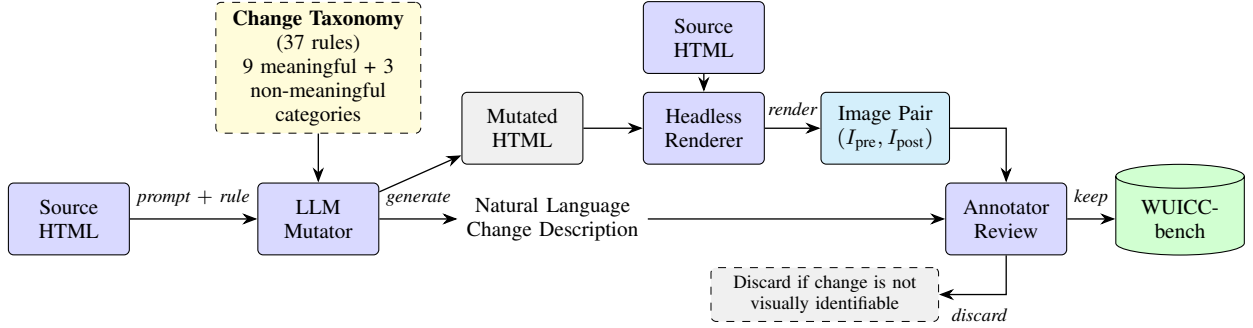
\begin{figure*}[t]
  \centering
  \vspace{-3mm}
  \begin{tikzpicture}[
    font=\footnotesize,
    every node/.style={align=center},
    html/.style={rectangle, draw, rounded corners=2pt, minimum height=0.95cm,
                 minimum width=1.6cm, fill=gray!12,  line width=0.4pt},
    stage/.style={rectangle, draw, rounded corners=2pt, minimum height=0.95cm,
                 minimum width=1.6cm, fill=blue!15,  line width=0.4pt},
    llm/.style={rectangle, draw, rounded corners=2pt, minimum height=0.95cm,
                minimum width=1.6cm, fill=orange!22, line width=0.4pt},
    proc/.style={rectangle, draw, rounded corners=2pt, minimum height=0.95cm,
                 minimum width=1.6cm, fill=teal!18,  line width=0.4pt},
    img/.style={rectangle, draw, rounded corners=2pt, minimum height=0.95cm,
                minimum width=1.7cm, fill=cyan!15,   line width=0.4pt},
    capnode/.style={minimum height=0.95cm,
    minimum width=2.0cm},
    human/.style={ minimum height=0.75cm,
                  minimum width=1.4cm},
    data/.style={cylinder, draw, shape border rotate=90, aspect=0.25,
                 minimum height=1.2cm, minimum width=1.7cm, fill=green!18,
                 line width=0.4pt},
    aux/.style={rectangle, draw, dashed, rounded corners=2pt, fill=yellow!18,
                inner sep=3pt, line width=0.3pt},
    drop/.style={rectangle, draw, dashed, rounded corners=2pt, fill=gray!12,
                 inner sep=3pt, line width=0.3pt, font=\scriptsize},
    arr/.style={-{Stealth[length=2.0mm]}, line width=0.5pt},
    lbl/.style={font=\scriptsize\itshape, midway, above=1.5pt},
    lblb/.style={font=\scriptsize\itshape, midway, below=1.5pt}
  ]

  \node[stage] (src)  at (0,    0) {Source\\HTML};
  \node[stage]  (mut)  at (3.3,  0) {LLM\\Mutator};
  \node[capnode]  (capn) at (6.4,  0) {Natural Language\\Change Description};
  \node[stage](ver)  at (12.4, 0) {Annotator\\Review};
  \node[data] (ds)   at (14.7, 0) {WUICC-\\bench};

  \node[html] (mhtml)  at (6.0,  1.2) {Mutated\\HTML};
  \node[stage] (render) at (8.4,  1.2) {Headless\\Renderer};
  \node[img]  (pair)   at (10.8, 1.2) {Image Pair\\$(I_{\text{pre}},I_{\text{post}})$};
  \node[stage] (srcb)   at (8.4,  2.4) {Source\\HTML};

  \node[aux, text width=2.5cm] (tax) at (3.3, 2.0)
    {\textbf{Change Taxonomy} (37 rules)\\
     9 meaningful + 3 non-meaningful\\categories};

  \node[drop, text width=2.8cm] (disc) at (10.0, -1.0)
    {Discard if change is not\\visually identifiable};

  \draw[arr] (src) -- (mut) node[lbl]{prompt $+$ rule};
  \draw[arr] (tax) -- (mut);

  \draw[arr] (mut) -- (capn) node[lbl]{generate};

  \draw[arr] (mut) -- (mhtml);

  \draw[arr] (srcb) -- (render);

  \draw[arr] (mhtml) -- (render);
  \draw[arr] (render) -- (pair)  node[lbl]{render};

  \draw[arr] (pair.east) -| (ver.north);

  \draw[arr] (capn) -- (ver);

  \draw[arr] (ver) -- (ds) node[lbl]{keep};
  \draw[arr] (ver.south) |- (disc.east) node[lblb, pos=0.7]{discard};

  \end{tikzpicture}
  \vspace{-6pt}
  \caption{WUICC-bench data generation pipeline. A controlled LLM-driven mutator edits the source HTML by applying exactly one change sampled from a 37-rule taxonomy. The original and mutated HTML are rendered into an image pair and paired with a natural language caption generated by the LLM to describe the applied change. Human annotators discard any sample whose caption does not completely and accurately describe the change visible in the rendered image pair.}
  \vspace{-16pt}
  \label{fig:overview}
\end{figure*}

\subsection{Visual Regression Testing}
VRT pipelines re-render UI screenshots after a code change and compare each one against its approved baseline to detect unintended visual regressions, surfacing every flagged difference to a human reviewer for an approve-or-reject decision.

Industry tools have advanced the state of practice considerably~\cite{vrt_sqa_ml_2025}, with Applitools Eyes~\cite{applitools} performing context-aware comparison beyond pixel matching, Percy~\cite{percy} suppressing common rendering artifacts in continuous-integration pipelines, and Chromatic~\cite{chromatic} detecting regressions at the component granularity. Yet none produces a natural language description of the detected change. WUICC produces exactly such a description, returning a flagged change in words rather than as a raw visual difference for the tester to interpret.
\subsection{Image Difference Captioning}
IDC generates a natural language description of the semantic differences between a pair of related images. A model must localize the changed regions, infer the semantic nature of each change, and verbalize it as a fluent sentence. Two domains have driven most prior research, natural image and remote sensing. In the natural image domain, the inputs are photographs of the same scene under different conditions. In the remote sensing domain, known as Remote Sensing Image Change Captioning (RSICC), the inputs are satellite or aerial images captured at two different time points.

Both settings share a two-stage formulation in which a visual stage extracts change-aware features from the image pair and a language stage decodes these features into a caption~\cite{chang2023changes,liu2022remote}. Web UI screenshots, however, differ substantially from both in visual statistics, layout regularity, text density, and the notion of a meaningful ``change'', raising the question of whether these methods transfer to the Web UI domain, which our benchmark is designed to answer.

\begin{table*}[t]
\centering
\caption{Taxonomy of Web UI changes used to construct WUICC-bench. Each sample contains exactly one atomic UI modification.}
\vspace{-6pt}
\label{tab:taxonomy}

\scriptsize
\setlength{\tabcolsep}{3pt}
\renewcommand{\arraystretch}{0.85}
\setlength{\aboverulesep}{1pt}
\setlength{\belowrulesep}{1pt}
\begin{tabular}{
>{\centering\arraybackslash}m{0.05\linewidth}
m{0.38\linewidth}
>{\centering\arraybackslash}m{0.05\linewidth}
m{0.35\linewidth}
}
\toprule

\multicolumn{4}{c}{\textbf{\textit{Meaningful Changes}}}
\\
\midrule

\multicolumn{2}{l}{\cellcolor{gray!15}\textbf{Missing elements}} & \multicolumn{2}{l}{\cellcolor{gray!15}\textbf{Resizing}}\\
No.~1 & Remove a button. & No.~15 & Enlarge or shrink existing elements such as buttons or cards.\\
No.~2 & Remove an input field or form. & \multicolumn{2}{l}{\cellcolor{gray!15}\textbf{Content update}}\\
No.~3 & Remove a navigation item. & No.~16 & Update displayed values such as prices, counts, or statistics.\\
No.~4 & Remove a section or module (e.g., header, footer, card block). & \multicolumn{2}{l}{\cellcolor{gray!15}\textbf{Thematic changes}}\\
\multicolumn{2}{l}{\cellcolor{gray!15}\textbf{Adding elements}} & No.~17 & Switch between light and dark themes.\\
No.~5 & Add a button with visible text. & No.~18 & Apply major color-scheme changes.\\
No.~6 & Add an input field or form. & \multicolumn{2}{l}{\cellcolor{gray!15}\textbf{Replacement}}\\
No.~7 & Add a navigation item. & No.~19 & Replace a button with a dropdown menu.\\
No.~8 & Add a new section or module. & No.~20 & Replace a button with a toggle switch.\\
\multicolumn{2}{l}{\cellcolor{gray!15}\textbf{Attribute modification}} & No.~21 & Replace label text with a tooltip or hint.\\
No.~9 & Modify visible text such as titles, labels, paragraphs, or placeholders. & No.~22 & Replace a paragraph with a card or collapsible panel.\\
\multicolumn{2}{l}{\cellcolor{gray!15}\textbf{Layout changes}} & No.~23 & Replace radio buttons with a dropdown selector.\\
No.~10 & Move an element to another interface region. & No.~24 & Replace an input field with a text area.\\
No.~11 & Change element alignment. & No.~25 & Replace a checkbox with a toggle switch.\\
No.~12 & Modify grid or flex layout structure. & No.~26 & Replace a submit button with a loading-state button.\\
\multicolumn{2}{l}{\cellcolor{gray!15}\textbf{Reordering}} & No.~27 & Replace a list with a card-grid layout.\\
No.~13 & Reorder menu or navigation items. & No.~28 & Replace a table with a sortable/filterable table.\\
No.~14 & Reorder lists or cards while preserving spacing. & No.~29 & Replace a static module with a collapsible panel.\\
 &  & No.~30 & Replace fixed navigation with bottom navigation.\\

\midrule

\multicolumn{4}{c}{\textbf{\textit{Non-meaningful changes}}}
\\

\midrule

\multicolumn{2}{l}{\cellcolor{gray!15}\textbf{Geometric shifts}}
&
\multicolumn{2}{l}{\cellcolor{gray!15}\textbf{Pure style changes}}
\\

No.~31 & Apply small positional shifts.
&
No.~35 & Apply small color adjustments.
\\

No.~32 & Apply minor spacing or padding adjustments.
&
No.~36 & Apply subtle font-size or font-weight changes.
\\

No.~33 & Apply subtle alignment changes.
&
No.~37 & Apply subtle typography variations.
\\

\multicolumn{2}{l}{\cellcolor{gray!15}\textbf{Dynamic content changes}}
&
&
\\

No.~34 &
Update transient content such as usernames, dates, or counters.
&
&
\\

\bottomrule
\end{tabular}
\vspace{-20pt}
\end{table*}
\subsection{Challenges in WUICC}
The Web UI setting introduces difficulties absent from prior IDC domains. We identify four core challenges that shape the difficulty of WUICC.
\begin{enumerate}
    \item \textbf{Layout diversity.} Web UI layouts follow no single visual pattern, varying greatly across pages in structure, element arrangement, and visual style. Combined with the wide variety of change types covered by the taxonomy, a model must handle an extremely large space of layout-change combinations with no fixed template to rely on, making generalization substantially harder than in domains where the visual background is more consistent.
    \item \textbf{Verbatim text reproduction.} A substantial fraction of UI changes are textual, affecting labels, headings, or body text. Unlike object-level changes in natural scenes, where paraphrasing is acceptable~\cite{jhamtani2018learning}, text in a UI must be transcribed verbatim. For example, replacing ``Submit'' with ``Send'' is a meaningful change, and the caption must reproduce the exact text ``Send''. Any other word, however semantically similar, constitutes an incorrect description, pushing the task beyond conventional captioning toward reading the rendered text and reproducing the changed content without any error.
    \item \textbf{High-resolution images with fine-grained changes.} Web UI screenshots are typically high-resolution, and a meaningful change often spans only a small fraction of the canvas. Detecting and describing changes at such granularity demands finer spatial sensitivity than the coarser changes typical of prior natural image IDC~\cite{jhamtani2018learning} and RSICC~\cite{liu2022remote,hoxha2022change} domains, where the altered content often occupies a large salient region of the scene.
    \item \textbf{Non-meaningful change suppression.} The WUICC setting includes non-meaningful modifications rooted in web rendering characteristics, such as sub-pixel shifts, minor spacing adjustments, and subtle style variations, which are defined by VRT practice as visual noise that should not be reported. Unlike noise in natural image or remote sensing settings, which arises from environmental factors such as lighting or seasonal variation~\cite{cheng2024change}, web rendering noise stems from deterministic and well-documented rendering mechanisms such as anti-aliasing, sub-pixel positioning, and font rasterization. Industry VRT tools attempt to suppress such noise~\cite{applitools,percy,chromatic}, but how effectively they do so is not publicly evaluated. A change captioning model, by contrast, must learn to suppress these artifacts and report ``no change'' regardless of whether the modification is visually detectable.
\end{enumerate}
\section{The WUICC-bench}\label{sec:dataset}
WUICC-bench is constructed through a controlled LLM-driven mutation pipeline, illustrated in Figure~\ref{fig:overview}. The pipeline is organized around a change taxonomy that defines what counts as a meaningful or non-meaningful UI modification and that governs both what changes are generated and how they are described. Because the taxonomy is the central design artifact, we present its rationale and categories first, then describe the generation procedure, human verification, and the source corpus from which the data is drawn.
\subsection{Design Rationale for the Change Taxonomy}\label{sec:taxonomy-rationale}

\textbf{Criteria.} A good taxonomy should meet certain design criteria, but no universally agreed set of them exists, so we identify the three we consider essential for WUICC. First, \emph{coverage} requires that the meaningful categories span the regression types that VRT tools and human testers are designed to surface, so the dataset does not systematically omit practically relevant changes. Second, \emph{separability} requires that the boundary between meaningful and non-meaningful changes and between different meaningful rules be operationalizable by the LLM at generation and by humans at audit, otherwise samples become ambiguous and verification collapses. Third, \emph{single-label composition} requires that each sample contain exactly one change. It keeps evaluation clean, since a model's success or failure on a sample can be traced to one specific change rather than to a mix of changes, and it keeps human verification tractable, since an annotator only needs to check that the caption completely and accurately describes a single change rather than disentangle several simultaneous edits.

\textbf{Sources informing the taxonomy.} To make the taxonomy as exhaustive as possible rather than an arbitrary set, we draw candidate Web UI change types both from repeatedly prompting an LLM to propose further changes until it surfaces no substantially different ones and from the changes documented in established VRT literature~\cite{adachi2018reducing,tanno2020region,aridome2026mixvrt,aridome2025prototype} and in practitioner accounts of industry VRT practice~\cite{vrt_sqa_ml_2025}. We then consolidate the combined set into mutually exclusive categories that reflect the changes VRT actually encounters.
\textbf{Design choices.} To keep the categories clearly separable and aligned with how regressions appear in practice, we also made the following design choices. First, we split element \emph{removal} from \emph{addition} rather than collapsing them into a single category. Second, we treat \emph{thematic changes} as meaningful but \emph{pure style changes} as non-meaningful. The difference is global-intentional, such as a light-to-dark mode switch, versus local-cosmetic, such as a single shadow tweak, mirroring how industry tools separate theme regressions from cosmetic style noise. Third, although \emph{layout changes} and \emph{resizing} are both meaningful and both spatial, we keep them as separate categories because their visual signatures differ, with layout changes affecting inter-element relationships and resizing affecting intra-element dimensions.

\textbf{Encoding.} We turn each criterion into a concrete prompt constraint. To serve coverage, the taxonomy is realized as a set of numbered atomic rules, so generation spans the full range of changes a Web UI may undergo. Separability is served at two boundaries. Between the meaningful and non-meaningful sides, every non-meaningful sample reuses a description drawn verbatim from a fixed phrase list, which stops the LLM from inventing spurious differences and keeps the two sides apart. Between individual rules, the atomic and mutually exclusive rule definitions let an auditor assign each sample to exactly one rule, and because meaningful changes are described by what is visible on the page rather than by markup-level selectors, that audit can be done on the rendered screenshot without inspecting the HTML. To serve single-label composition, the LLM is instructed to apply exactly one specified rule per sample, making the taxonomy single-label by construction. Compound regressions are out of scope of our work and recorded as a limitation in Section~\ref{sec:limitations}.

Table~\ref{tab:taxonomy} summarizes the twelve categories encoded in the prompt. Nine are meaningful categories, whose changes must be described in the generated caption, and three are non-meaningful categories, whose samples carry a fixed ``no change'' description.
\subsection{LLM-Driven Sample Generation}\label{sec:generation}
Figure~\ref{fig:overview} summarizes the pipeline that turns a source page into a verified sample through four stages.

\textbf{(1) Source HTML.} Each sample starts from a source HTML page and the change taxonomy of 37 atomic rules. For every sample we draw one rule uniformly from the taxonomy and feed it to the mutator alongside the page.

\textbf{(2) LLM Mutator.} Prompted with the page HTML and the selected rule, the LLM applies that single change and, in one pass, emits both the mutated HTML and a natural language description of the change. Producing the edit and its description together keeps them aligned by construction, avoiding the drift a separate captioning pass would introduce. For a non-meaningful rule the description reuses a phrase drawn from a small fixed list of interchangeable no-change phrases, which stops the model from inventing spurious differences.

\textbf{(3) Headless Renderer.} The original and the mutated HTML are rendered under an identical viewport, producing images $I_{\text{pre}}$ and  $I_{\text{post}}$, which  differ only in the intended change.

\textbf{(4) Annotator Review.} Annotators review the candidate pair against its description and either keep the sample for WUICC-bench or discard it. Section~\ref{sec:human-verification} details the verification stage and the failure modes it targets.

Because the pipeline is driven entirely by the prompt and indexed by atomic rules, it scales to large corpora without manual authoring of changes. Both the taxonomy and the rule-indexed pipeline are thus reusable beyond WUICC-bench, supporting extension to new change types and the construction of similar benchmarks in other structured-UI domains.
\begin{table}[t]
\caption{Distribution of screenshot resolutions in WUICC-bench.}
\vspace{-6pt}
\label{tab:resolution}
\centering
\footnotesize
\renewcommand{\arraystretch}{0.9}
\setlength{\tabcolsep}{4pt}
\setlength{\aboverulesep}{1pt}
\setlength{\belowrulesep}{1pt}
\resizebox{\linewidth}{!}{%
\begin{tabular}{>{\bfseries}lrr@{\hspace{1em}}>{\bfseries}lrr}
\toprule
\textbf{Resolution} & \textbf{Count} & \textbf{\%} & \textbf{Resolution} & \textbf{Count} & \textbf{\%} \\
\midrule
$1280\times720$ & 12,949 & 65.4 & $1280\times[1500,2000)$ & 336 & 1.7 \\
$1280\times[721,800)$ & 1,223 & 6.2 & $1280\times[2000,4320]$ & 107 & 0.5 \\
$1280\times[800,1000)$ & 2,118 & 10.7 & $>1280$ (outliers) & 66 & 0.3 \\
$1280\times[1000,1200)$ & 1,772 & 8.9 & Total & 19,812 & 100.0 \\
$1280\times[1200,1500)$ & 1,241 & 6.3 & & & \\
\bottomrule
\end{tabular}}
\vspace{-20pt}
\end{table}

\subsection{Human Verification}\label{sec:human-verification}
Generated samples inevitably contain failure cases, and a human verification stage is therefore introduced to filter them out. Through manual inspection of a large pool of generated samples, we identify three recurring failure modes. In the first, the LLM fails to apply the intended edit to the HTML, so the rendered image pair shows no change while the description still claims one. In the second and most common, the LLM applies the edit and its textual description is consistent with it, but the rendered image pair still fails to reflect the intended change. In the third, the textual description does not faithfully describe the HTML edit, so although the image pair is correctly rendered from the edit, the description does not state the change it actually shows. To address both, each sample is reviewed by one of two annotators, given only the image pair and the description, both of whom were trained on the change taxonomy and the review criterion before annotation. A sample is retained only if its caption completely and accurately describes the change visible in the rendered pair, and discarded otherwise.

\subsection{LLM Selection}
We use GPT-5.1~\cite{gpt51} as the mutator for its established strong performance on challenging reasoning and code-related tasks, which our pipeline relies on to apply a structural HTML edit and describe the resulting change accurately. Among the available options, the larger GPT-5.5 is more capable but substantially more expensive, whereas the smaller GPT-5-mini is cheaper but a weaker model, so GPT-5.1 offers a reasonable balance of capability and cost. Because the pipeline conditions only on HTML and never incurs vision tokens, the resulting cost is low, making large-scale construction economical.
\subsection{Dataset Statistics and Comparison}
WUICC-bench is built from the 37 atomic rules of our change taxonomy. Each of these rules is sampled with equal probability during generation, giving an intended uniform distribution over rules. For each atomic rule, only a small fraction of generated samples are invalid, so the distribution is largely preserved. The final dataset retains 9,906 valid samples, of which 8,583 contain meaningful changes and 1,323 contain non-meaningful changes.

Table~\ref{tab:resolution} summarizes the resolution distribution of the screenshots, where each sample contributes a before and after screenshot and each row reports a resolution group (width $\times$ height) with the height binned into intervals. All pages are rendered at a fixed viewport width of 1280 pixels, and about two thirds of the screenshots match the standard $1280\times720$ viewport. The remainder are full-page captures whose height grows with page content, extending to 4320 pixels, with only 0.3\% of screenshots wider than 1280 pixels. The screenshots are thus consistently high-resolution, so a single change occupies only a small fraction of the canvas, realizing the challenge of fine-grained changes identified in Section~\ref{sec:background}.

Figure~\ref{fig:desclen} shows the distribution of valid word counts per caption. The sharp peak at four words corresponds to the fixed short phrases attached to non-meaningful samples, whereas the captions of meaningful changes form a broad distribution that peaks around 18 words and decays slowly, with 102 captions exceeding 50 words. Models must therefore handle both terse ``no change'' outputs and long, detailed change descriptions.
\begin{figure}[t]
\centering
\vspace{-3mm}
\resizebox{\linewidth}{!}{%
\begin{tikzpicture}
\begin{axis}[
  width=\linewidth, height=4cm,
  ybar, bar width=3pt,
  axis lines*=left,
  xlabel={Valid word counts in each caption},
  xlabel near ticks,
  xlabel style={yshift=5pt},
  ytick=\empty,
  clip=false,
  xmin=1, xmax=53, ymin=0, ymax=730,
  xtick={5,10,15,20,25,30,35,40,45,51},
  xticklabels={5,10,15,20,25,30,35,40,45,$>$50},
  tick align=outside,
  label style={font=\footnotesize},
  tick label style={font=\scriptsize},
]
\addplot[fill=blue!60, draw=none, bar shift=0pt] coordinates {
(3,26)(6,139)(7,7)(8,145)(9,89)(10,153)(11,238)(12,243)(13,215)
(14,311)(15,443)(16,447)(17,525)(18,644)(19,611)(20,510)(21,472)(22,430)
(23,375)(24,339)(25,289)(26,251)(27,251)(28,202)(29,183)(30,173)(31,164)
(32,148)(33,101)(34,87)(35,78)(36,67)(37,42)(38,34)(39,32)(40,32)(41,30)
(42,23)(43,22)(44,15)(45,10)(46,11)(47,7)(48,15)(49,10)(50,6)
};
\addplot[fill=blue!60, draw=none, bar shift=0pt] coordinates {(4,700)};
\addplot[fill=blue!30, draw=none, bar shift=0pt] coordinates {(51,102)};
\draw[white, line width=1.6pt] (axis cs:3.55,640) -- (axis cs:4.45,660);
\draw[blue!60, line width=0.4pt] (axis cs:3.45,633) -- (axis cs:4.55,653);
\draw[blue!60, line width=0.4pt] (axis cs:3.45,645) -- (axis cs:4.55,665);
\node[font=\scriptsize, anchor=south] at (axis cs:4,700) {1159};
\node[font=\scriptsize, anchor=south] at (axis cs:18,644) {644};
\node[font=\scriptsize, anchor=south] at (axis cs:51,102) {102};
\end{axis}
\end{tikzpicture}}
\vspace{-20pt}
\caption{Distribution of valid word counts per caption in WUICC-bench. Words beyond 50 are aggregated, and the four-word bar is truncated for readability.}
\vspace{-20pt}
\label{fig:desclen}
\end{figure}
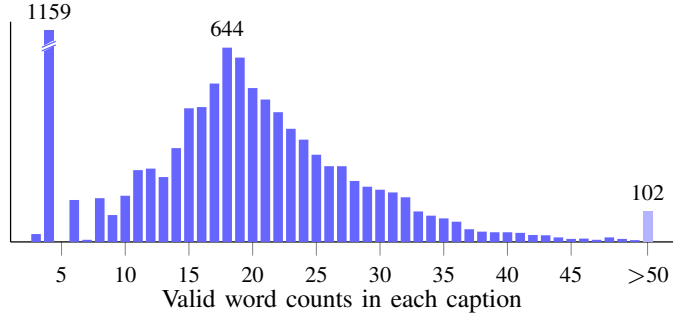

\begin{figure*}[t]
\centering
\vspace{-3pt}
\footnotesize
\newcommand{\dcell}[1]{\tikz[baseline=(dc.north)]{%
  \node[draw=gray!65, dashed, dash pattern=on 2pt off 1.5pt, line width=0.5pt,
        rounded corners=2pt, inner sep=3pt, text width=0.48\textwidth,
        align=left, anchor=north] (dc) {#1};}}
{\bfseries Representative WUICC-bench captions for changes of the same type: addition}\par\vspace{0.0mm}
\setlength{\tabcolsep}{4pt}
\renewcommand{\arraystretch}{0.9}
\setlength{\lineskip}{0pt}
\begin{tabular}{@{}c@{\hspace{3pt}}c@{}}
\dcell{\textbf{(a)} A new navigation item labeled ``Search'' was added as a section at the bottom of the main content with a simple search form.} &
\dcell{\textbf{(b)} A new information section ``Contact Us'' with address, email, and phone text was added where the contact form placeholder was.}\\[-4pt]
\dcell{\textbf{(c)} A new dropdown labeled ``Preferred Contact Method'' was added under the main welcome text, replacing the need for separate radio buttons.} &
\dcell{\textbf{(d)} A new section titled ``Key Statistics'' was added at the end of the main content, showing example numbers for programs, faculty, and students.}\\[-4pt]
\dcell{\textbf{(e)} The original header navigation remains at the top, and a new bottom navigation bar with the same items plus a floating-style ``Donate Now'' button has been added to the footer, turning navigation into a bottom bar with a prominent action button.} &
\dcell{\textbf{(f)} In the first section under the text ``It is a section of the website...'', a new message area was added consisting of a label ``Your Message'' and a multiline text box with placeholder ``Write your message here'', replacing the need for a single-line input field with a text area.}\\
\end{tabular}
\vspace{-1mm}
\caption{Illustration of caption variability in WUICC-bench.}
\vspace{-7mm}
\label{fig:capdiv}
\end{figure*}

We then split the dataset into training, validation, and test sets with a ratio of approximately 70:10:20, giving 6,963 training, 982 validation, and 1,961 test samples.
Two properties of the resulting dataset are worth highlighting, described below.

\noindent \textbf{a) Caption style.} Because descriptions are produced by an LLM rather than written by a specific or handful human annotators, they exhibit substantial lexical and syntactic variability, reaching up to 113 valid words. They are also considerably longer than typical captions in the natural image or remote sensing settings. The same type of change can be described in multiple valid phrasings, as shown in Fig.~\ref{fig:capdiv}, more faithfully reflecting how human testers would phrase the same regression in different reports.

\noindent \textbf{b) Scale.} WUICC-bench contains 9{,}906 samples, substantially larger than change captioning datasets such as Dubai-CC~\cite{hoxha2022change} (500 pairs) and Image Editing Request~\cite{tan2019expressing} (3{,}939 pairs), and comparable in size to larger human-annotated benchmarks such as Spot-the-Diff~\cite{jhamtani2018learning} (13{,}192 pairs) and LEVIR-CC~\cite{liu2022remote} (10{,}077 pairs), a scale sufficient to support benchmarking.

\noindent \textbf{c) Distinguishing properties.} Beyond scale, WUICC-bench differs from prior datasets on two axes that matter for VRT, namely its programmatically rendered, pixel-aligned, viewpoint-free Web UI domain, and its verbatim-text requirement under which a textual edit must be transcribed exactly rather than paraphrased.
\subsection{Source HTML Corpus}\label{sec:source}
Having characterized the resulting dataset, we now turn to the source corpus its pages are drawn from and the rationale behind it. The source HTML is drawn from WebSight~\cite{laurenccon2024unlocking}, which spans a wide range of Web UI structures and styles. WebSight consists of LLM-synthesized pages designed to resemble real Web UIs. In its released form, the corpus is self-contained static HTML that renders deterministically, which is essential for a benchmark in which each pre/post pair must differ only in the intended modification.

We build on the released corpus rather than scraping production sites ourselves so that the resulting benchmark can be openly redistributed. Scraped production HTML is typically not redistributable under copyright and terms-of-service constraints, whereas WebSight is released under terms that permit us to re-render, mutate, and release the derived artifacts. The residual gap to live production traffic is acknowledged in Section~\ref{sec:limitations}.

From the source corpus, we select pages that are complex along two dimensions, layout and interaction. Layout complexity is measured by the number of top-level UI blocks, and interaction complexity by the number of clickable elements such as buttons. These criteria filter out overly simple pages and retain ones with multi-column or grid-based layouts, nested structures, and multiple functional regions, which offer rich potential for diverse modifications.
\subsection{Task Formulation}
Having described how WUICC-bench is constructed, we now formalize the task it is designed to support, which applies IDC to the Web UI domain. Given an ordered pair of Web UI screenshots, a baseline $I_{\text{pre}}$ and a candidate $I_{\text{post}}$ rendered from the original and the modified page, the goal is to learn a mapping $f: (I_{\text{pre}}, I_{\text{post}}) \rightarrow Y$ that generates a natural language caption $Y=\{y_1,\ldots,y_T\}$ with $y_t \in \mathcal{V}$ describing the change from $I_{\text{pre}}$ to $I_{\text{post}}$, where $\mathcal{V}$ is the vocabulary and $T$ the caption length. The formulation turns the output of regression checking from a semantically blind difference signal into a natural language description, so that a tester is told what changed in words instead of being handed a highlighted region to interpret. It also makes the substantial body of IDC methods developed for natural images and remote sensing directly applicable to VRT for the first time, enabling a concrete study of whether their modeling paradigms transfer to the Web UI setting. Models are trained by maximizing the likelihood of the ground-truth caption under a standard token-level cross-entropy loss:
  \[
  \mathcal{L} = - \sum_{t=1}^{T} \log p(y_t \mid I_{\text{pre}}, I_{\text{post}}, y_{<t}),
  \tag{1}\label{eq:ce}
  \]
where $p(.)$ denotes probability. While some prior works augment the objective with auxiliary losses, such as a cross-view contrastive loss in SCORER~\cite{tu2023self} or attention and entropy regularization in VARD~\cite{tu2023adaptive}, Eq.~(\ref{eq:ce}) remains the fundamental training signal common to virtually all change captioning models~\cite{chang2023changes,zhou2024single,cai2023interactive,liu2022remote,liu2024rscama,sun2024lightweight,xian2025dynamic,cao2026rmnet}.
\section{Experiments}\label{sec:experiments}
\subsection{Research Questions}
Our experiments aim to answer the following important research questions (RQs):
\begin{itemize}[leftmargin=10pt]
    \item \textbf{RQ1: Does our LLM-driven generation pipeline produce a reliable benchmark, i.e., do the mutated HTML and its paired change caption yield valid samples that human annotators judge consistently?}

    To answer it, we measure the inter-annotator agreement and the accuracy of the LLM-driven generation pipeline, requiring the mutated HTML to render the intended change and the paired caption to describe it correctly.

    \item \textbf{RQ2: How well do existing IDC methods, originally developed for remote sensing and natural images, and general-purpose LLMs perform on WUICC?}

    To answer it, we evaluate eleven representative IDC methods and two zero-shot general-purpose LLMs on WUICC-bench.

    \item \textbf{RQ3: Can the benchmarked methods suppress non-meaningful changes and report ``no change'', avoiding the false positives that pixel-level comparison produces in VRT?}

    To answer it, we measure the no-change and change accuracy on the non-meaningful and meaningful categories, respectively, and compare against pixel-level baselines.

    \item \textbf{RQ4: How do key configuration factors, namely input image resolution and beam size, affect performance?}

    To answer it, we run ablation studies on input resolution and beam size with representative methods.
\end{itemize}
\subsection{Benchmark Methods}
We benchmark eleven representative methods spanning RSICC and natural IDC, all re-implemented and trained on WUICC-bench. Each re-implementation follows the architecture and hyperparameters reported in the original work and builds on the authors' released code where it is available. We release all re-implementations and configurations so that their correctness can be independently verified. In addition to these trained methods, we evaluate two general-purpose VLMs, Llama-3.2-11B-Vision-Instruct~\cite{grattafiori2024llama} and Qwen2-VL-7B-Instruct~\cite{wang2024qwen2}, in a zero-shot setting as reference baselines. We summarize each trained method below.

\begin{itemize}[leftmargin=10pt]
    \item \textbf{Chg2Cap}~\cite{chang2023changes}. Siamese CNN backbone with hierarchical self-attention encoder and Transformer decoder.
    \item \textbf{SEN}~\cite{zhou2024single}. Contrastive-pretrained single-stream visual extractor with a difference encoder guided by cross-attention.
    \item \textbf{ICT-Net}~\cite{cai2023interactive}. Multi-scale CNN features with interactive change-aware encoder and cross gated-attention decoder.
    \item \textbf{RSICCformer}~\cite{liu2022remote}. CNN with dual-branch Transformer encoder and difference-aware cross-attention.
    \item \textbf{RSCaMa}~\cite{liu2024rscama}. CLIP image encoder with stacked Mamba CaMa layers and Transformer decoder.
    \item \textbf{SparseFocus}~\cite{sun2024lightweight}. ResNet-101 backbone with sparse focus attention encoder and Transformer decoder.
    \item \textbf{DAE}~\cite{xian2025dynamic}. Dynamic asymmetric encoder via difference-guided dynamic convolution and temporal modeling.
    \item \textbf{RMNet}~\cite{cao2026rmnet}. Dual-stream CNN--VMamba network with difference recalibration to suppress pseudo-changes.
    \item \textbf{CARD}~\cite{tu2024context}. Decomposition of image-pair features into shared and change-specific components under consistency and independence constraints.
    \item \textbf{VARD}~\cite{tu2023adaptive}. Context- and position-aware representations with change isolation based on feature subtraction.
    \item \textbf{SCORER}~\cite{tu2023self}. Contrastive view-invariant representations with cross-modal backward reasoning.
\end{itemize}

Despite their architectural differences, these methods share a common structure. Each pairs a feature extraction backbone, a CNN such as ResNet-101 or a vision Transformer, with a change-aware encoder that models the differences between the image pair and a Transformer-based caption decoder. The change-aware encoder, which sets change captioning apart from single-image captioning, is where the methods differ most.

The two VLM baselines are evaluated zero-shot, with no training on WUICC-bench. Each receives the before and after screenshots together with a fixed instruction to describe the change in one caption, and its output is scored under the same metrics and ``no change'' convention as the trained methods. They establish how far an off-the-shelf VLM reaches on WUICC without any domain adaptation.
\subsection{Evaluation Metrics}
Traditional VRT evaluation is detection-oriented, measuring regression-versus-noise classification or the reduction in differences a tester must confirm~\cite{adachi2018reducing,tanno2020region,adachi2020method}. Such metrics score a binary flag or a highlighted region and cannot assess the content of a generated description. We therefore adopt the standard language-level metrics from IDC, which give a reproducible and widely used basis for comparison with prior work. Designing UI-specific metrics that capture verbatim text fidelity is an important direction for future work.

\begin{table}[t]
\centering
\caption{Inter-annotator agreement results on the sampled data.}
\vspace{-6pt}
\label{tab:kappa}
\begin{tabular}{>{\bfseries}lc}
\toprule
\textbf{Metric} & \textbf{Result} \\
\midrule
Cohen's Kappa ($\kappa$) & 0.722 \\
\bottomrule
\end{tabular}
\vspace{-18pt}
\end{table}

We report the standard language-level captioning metrics, namely BLEU-n (n=1,2,3,4)~\cite{papineni2002bleu}, METEOR~\cite{banerjee2005meteor}, ROUGE\_L~\cite{chin2004rouge}, CIDEr~\cite{vedantam2015cider}, and SPICE~\cite{anderson2016spice}. They differ in the kind of overlap they reward. BLEU scores n-gram precision and ROUGE\_L scores longest-common-subsequence overlap, METEOR adds stem, synonym, and paraphrase matching, CIDEr weights n-grams by TF-IDF so that distinctive content counts more, and SPICE compares scene-graph tuples of objects, attributes, and relations.
\subsection{Implementation Details}
Unless stated otherwise, each before/after screenshot is resized to $512\times512$. The baselines were originally designed for $256\times256$ inputs, which is too coarse to resolve the dense text and fine-grained changes in WUICC-bench. We therefore raise the resolution, but adopt $512\times512$ rather than a higher resolution, since it captures the detail that $256\times256$ misses while remaining computationally tractable for every baseline, unlike $768\times768$, which our ablation (Table~\ref{tab:res-ablation}) finds most accurate but contains $2.25\times$ as many pixels, raising compute and memory cost accordingly. We further examine the effect of input resolution in the ablation study. The captions are truncated to at most 50 valid words, which affects only the 102 captions (about 1\% of the dataset) that are longer. Each baseline keeps its original architecture and published hyperparameters, including optimizer, learning rate, and batch size, while sharing the same data pipeline, tokenization, and metric implementation. Each method is optimized with its original training objective from the corresponding paper, namely the cross-entropy loss of Eq.~\ref{eq:ce} together with any method-specific auxiliary losses, on a single NVIDIA H100 GPU. At inference, captions are generated by beam search with a width of 3, and all reported numbers are computed on the test split.
\subsection{Quantitative Results}
\subsubsection{Agreement Rate}
To assess benchmark reliability, we first present the results of human verification. We select 100 samples for evaluation, deliberately over-sampling invalid generations so that valid and invalid cases are both well represented, which avoids the inflated chance agreement that a heavily skewed sample would produce. Each sample is independently annotated by two annotators, and inter-annotator reliability is measured with Cohen's Kappa. The two annotators judge 48 and 60 of the 100 samples as valid, respectively, and agree on 86 samples, with 47 judged valid and 39 judged invalid by both. As reported in Table~\ref{tab:kappa}, the result is $\kappa \approx 0.722$, which corresponds to substantial agreement~\cite{landis1977measurement} and suggests that the annotations are reliable.
\subsubsection{LLM Accuracy}
We then present the accuracy of the LLM-driven generation, computed over the entire pool of generated samples rather than the 100-sample agreement subset, and defined as the proportion that the verification stage of Section~\ref{sec:human-verification} retains as correct. The failure cases fall into three types. First, the LLM fails to modify the code, which is attributed to the limitations of the LLM itself. Second, the LLM successfully modifies the code, but the rendered UI image does not reflect the intended change, which is attributed to the UI rendering process. Third, the LLM successfully modifies the code, but the generated caption fails to accurately describe the change, which is attributed to the captioning process. As reported in Table~\ref{tab:acc}, the pipeline retains 78.09\% of generated samples as correct, which indicates its effectiveness in automating the task and reducing the manual annotation cost. Notably, the 78.09\% and the agreement coefficient $\kappa$ capture different aspects of reliability. The acceptance rate measures how often the pipeline yields a correct sample over the full pool, while $\kappa$ measures how consistently annotators apply the verification criterion on the 100-sample subset. The two are therefore complementary rather than measurements of the same quantity.
\begin{table}[t]
\centering
\caption{Overall accuracy of the LLM-driven generation.}
\vspace{-6pt}
\label{tab:acc}
\begin{tabular}{>{\bfseries}lc}
\toprule
\textbf{Metric} & \textbf{Result (\%)} \\
\midrule
Accuracy & 78.09 \\
\bottomrule
\end{tabular}
\vspace{-18pt}
\end{table}

\rqbox{\textbf{Answer to RQ1.} The two annotators reach substantial agreement and the pipeline attains high generation accuracy, with the mutated HTML reliably rendering the intended change and the paired caption describing it, indicating that the constructed benchmark is reliable enough for evaluation.}
\subsubsection{Results on WUICC-bench}
With the benchmark validated, we now examine method performance by comparing the eleven trained IDC methods on WUICC-bench, together with the two zero-shot VLM baselines, Llama-3.2-11B-Vision-Instruct and Qwen2-VL-7B-Instruct. Table~\ref{tab:bench} reports the comparison, and among the trained methods no single one dominates. SparseFocus~\cite{sun2024lightweight} attains the best n-gram and CIDEr scores, reaching 0.2571 BLEU-4 and 2.0761 CIDEr, with Chg2Cap~\cite{chang2023changes} close behind. On SPICE the ordering flips, as Chg2Cap is best at 0.3345 while SparseFocus falls to near the bottom at 0.2579. The divergence is itself informative, because it shows that strong lexical overlap on n-gram measures does not guarantee semantically adequate captions, given that SPICE scores agreement over scene-graph propositions of objects, attributes, and relations rather than over surface word sequences. A second and orthogonal pattern is architectural rather than metric-specific, and it tracks the domain assumptions embedded in each method's change-aware encoder. The three methods originally engineered for natural-image IDC, namely the viewpoint-robust and disentanglement designs of SCORER~\cite{tu2023self}, VARD~\cite{tu2023adaptive}, and CARD~\cite{tu2024context}, all settle into the lower half of the ranking across both n-gram and SPICE metrics. Because Web UI screenshots are programmatically rendered and therefore pixel-aligned, free of the camera viewpoint shifts that these designs are built to neutralize, the additional invariance machinery they carry adds parameters and optimization burden without addressing any difficulty that the Web UI setting actually presents. The two zero-shot VLMs trail the trained methods on most metrics, with Llama-3.2-11B scoring only 0.0363 BLEU-4 and Qwen2-VL 0.0920, although Qwen2-VL reaches a competitive 1.8699 CIDEr, showing that a general-purpose VLM can capture distinctive changed content yet still falls short of the n-gram and SPICE fidelity the trained methods achieve. Overall the absolute scores remain low across both groups, indicating the difficulty of our benchmark for the trained-from-scratch IDC paradigm and for off-the-shelf VLMs alike.

\rqbox{\textbf{Answer to RQ2.} Existing IDC methods transfer poorly to WUICC. No single method dominates across metrics, and the absolute scores remain low, while two zero-shot VLM baselines fare worse still, suggesting that Web UI changes are challenging both for methods designed for remote sensing and natural images and for off-the-shelf VLMs.}
\begin{table}[t]
  \caption{Performance comparison on WUICC-bench. Higher is better. The \hlbest{best} and \hlsecond{second-best} results in each column are highlighted.}
  \vspace{-6pt}
  \label{tab:bench}
  \centering
  \resizebox{\linewidth}{!}{
  \begin{tabular}{>{\bfseries}lcccccccc}
    \toprule
    \rowcolor{cyan!25}
    \textbf{Method} & \textbf{BLEU-1} & \textbf{BLEU-2} & \textbf{BLEU-3} & \textbf{BLEU-4} & \textbf{METEOR} & \textbf{ROUGE\_L} & \textbf{CIDEr} & \textbf{SPICE}\\
    \midrule
    Chg2Cap & \hlsecond{0.4530} & \hlsecond{0.3585} & \hlsecond{0.2923} & \hlsecond{0.2429} & \hlsecond{0.2347} & \hlsecond{0.4876} & \hlsecond{1.9492} & \hlbest{0.3345} \\
    SEN & 0.4385 & 0.3399 & 0.2694 & 0.2163 & 0.2209 & 0.4609 & 1.6644 & 0.3184 \\
    ICT-Net & 0.4234 & 0.3261 & 0.2605 & 0.2129 & 0.2168 & 0.4532 & 1.7363 & 0.3022 \\
    RSICCformer & 0.3846 & 0.3000 & 0.2413 & 0.1985 & 0.2028 & 0.4347 & 1.6104 & 0.2992 \\
    RSCaMa & 0.4465 & 0.3441 & 0.2719 & 0.2194 & 0.2249 & 0.4745 & 1.7666 & \hlsecond{0.3185} \\
    SparseFocus & \hlbest{0.4725} & \hlbest{0.3779} & \hlbest{0.3093} & \hlbest{0.2571} & \hlbest{0.2438} & \hlbest{0.5060} & \hlbest{2.0761} & 0.2579 \\
    DAE & 0.3939 & 0.2957 & 0.2326 & 0.1887 & 0.2017 & 0.4296 & 1.5492 & 0.2911 \\
    RMNet & 0.3464 & 0.2578 & 0.2016 & 0.1626 & 0.1792 & 0.4000 & 1.4851 & 0.2667 \\
    CARD & 0.3860 & 0.2895 & 0.2270 & 0.1835 & 0.1954 & 0.4253 & 1.5657 & 0.2793 \\
    VARD & 0.4435 & 0.3288 & 0.2546 & 0.2035 & 0.2160 & 0.4177 & 1.1921 & 0.2786 \\
    SCORER & 0.3301 & 0.2314 & 0.1743 & 0.1373 & 0.1652 & 0.3679 & 1.2991 & 0.2403 \\
    Llama-3.2-11B-Vision-Instruct & 0.2081 & 0.1065 & 0.0570 & 0.0363 & 0.1057 & 0.2624 & 0.4860 & 0.0861 \\
    Qwen2-VL-7B-Instruct & 0.2290 & 0.1558 & 0.1158 & 0.0920 & 0.1518 & 0.3949 & 1.8699 & 0.2498 \\
    \bottomrule
  \end{tabular}
  }
\vspace{-4pt}
\end{table}
\begin{table}[t]
  \caption{Change-awareness accuracy (\%). Higher is better for both columns.}
  \vspace{-6pt}
  \label{tab:suppression}
  \centering
  \tiny
  \begin{tabular}{>{\bfseries}lcc}
    \toprule
    \rowcolor{cyan!25}
    \textbf{Method} & \textbf{No-Change Acc.} & \textbf{Change Acc.} \\
    \midrule
    Pixel Diff & 0.00 & 100.00 \\
    Pixel Diff (w/ tol.) & 6.72 & 97.93 \\
    \midrule
    SEN & 96.27 & 96.16 \\
    ICT-Net & 97.01 & 93.80 \\
    RSICCformer & 85.82 & 95.33 \\
    RMNet & 97.39 & 93.55 \\
    Llama-3.2-11B-Vision-Instruct & 21.64 & 79.39 \\
    Qwen2-VL-7B-Instruct & 99.63 & 65.68 \\
    \bottomrule
  \end{tabular}
\vspace{-16pt}
\end{table}
\subsubsection{Change Awareness}
A core requirement for VRT is false positive suppression, namely suppressing visual noise rather than reporting it, without becoming blind to genuine changes. We therefore report two accuracies in Table~\ref{tab:suppression}. ``No-Change Acc.'' is the suppression accuracy on non-meaningful samples, the proportion correctly described as ``no change''. ``Change Acc.'' is the proportion of meaningful samples correctly reported as changed rather than suppressed. Higher is better for both, and reporting them together guards against a degenerate model that trivially emits ``no change'' everywhere. The goal here is not to rank all methods on suppression, which Table~\ref{tab:bench} already covers, but to contrast the captioning paradigm with pixel-level comparison, so we report four IDC methods that span the architectural families in our benchmark, the CNN-based SEN and ICT-Net, the dual-branch Transformer RSICCformer, and the CNN-VMamba RMNet, together with the two zero-shot VLMs rather than all eleven IDC methods. As a reference we include a naive pixel-level comparison (``Pixel Diff'') and one that applies a 0.1 difference threshold with anti-aliasing tolerance (``Pixel Diff (w/ tol.)'').

The two pixel-diff references illustrate a known weakness of pixel-level comparison. Both flag almost every meaningful change, reaching a change accuracy of 100.00\% and 97.93\%, yet they suppress almost nothing, with a no-change accuracy of only 0.00\% and 6.72\%. Any sub-pixel shift, minor restyle, or dynamic-content update is therefore reported as a difference, the kind of false positive that VRT pipelines must triage by hand. Adding an anti-aliasing tolerance raises suppression by under seven points and already begins to miss genuine changes, suggesting that a single fixed threshold struggles to separate noise from signal.

The four trained IDC methods behave in the opposite and desirable way. They suppress the large majority of non-meaningful changes, with a no-change accuracy between 85.82\% and 97.39\%, while still reporting most genuine changes, with a change accuracy between 93.55\% and 96.16\%. RMNet and ICT-Net reach the highest suppression among the IDC methods at 97.39\% and 97.01\%, and none collapses to the degenerate solution of always emitting ``no change'', since every IDC change accuracy stays above 93\%, suggesting that these models capture change at a more semantic than pixel level. The two zero-shot VLMs behave far less stably. Qwen2-VL suppresses almost all non-meaningful changes, reaching 99.63\% no-change accuracy, but does so by under-reporting genuine ones, with its change accuracy falling to 65.68\%, whereas Llama-3.2-11B fails in the opposite direction with only 21.64\% no-change accuracy. Their change accuracies, 65.68\% and 79.39\%, both fall well below those of the trained methods, indicating that off-the-shelf VLMs do not yet provide the selective suppression VRT needs.

\rqbox{\textbf{Answer to RQ3.} The trained IDC methods suppress non-meaningful changes far more selectively than pixel-level comparison while still reporting almost all genuine changes, suggesting they capture change at a semantic rather than pixel level, the capability VRT needs to avoid false positives. Two zero-shot VLMs, by contrast, trade suppression against sensitivity and do not yet reach such a balance.}
\subsection{Ablation Studies}
We study the sensitivity of performance to two design factors, input image resolution and beam size.
\subsubsection{Influence of Input Image Resolution}
Since our screenshots are high-resolution, the input resolution fed to the models may affect their performance. We therefore study its impact by evaluating three settings, $256\times256$, $512\times512$, and $768\times768$, which probe the sensitivity of each model to visual granularity in Web UI change understanding. We run the ablation on five representative methods, with results reported in Table~\ref{tab:res-ablation}. Across all methods, the best scores consistently occur at $768\times768$, and $512\times512$ is usually second best, indicating that higher input resolution generally yields better performance, but higher resolution also requires more compute and memory, which is why we adopt $512\times512$ as the default rather than $768\times768$. 

\begin{table}[t]
  \caption{Effect of beam size on three representative methods. Higher is better. The \hlbest{best} and \hlsecond{second-best} results in each column are highlighted.}
  \vspace{-6pt}
  \label{tab:beam}
  \centering
  \resizebox{\linewidth}{!}{
  \begin{tabular}{>{\bfseries}cccccccccc}
    \toprule
    \rowcolor{cyan!25}
    \textbf{Method} & \textbf{Beam Size} & \textbf{BLEU-1} & \textbf{BLEU-2} & \textbf{BLEU-3} & \textbf{BLEU-4} & \textbf{METEOR} & \textbf{ROUGE\_L} & \textbf{CIDEr} & \textbf{SPICE}\\
    \midrule
    \multirow{3}{*}{Chg2Cap} & 1 & \hlbest{0.4739} & \hlbest{0.3711} & \hlbest{0.2993} & \hlbest{0.2458} & \hlbest{0.2381} & \hlbest{0.4897} & 1.9237 & 0.3303 \\
       & 3 & \hlsecond{0.4530} & \hlsecond{0.3585} & \hlsecond{0.2923} & \hlsecond{0.2429} & \hlsecond{0.2347} & \hlsecond{0.4876} & \hlbest{1.9492} & \hlbest{0.3345} \\
       & 5 & 0.4417 & 0.3500 & 0.2856 & 0.2376 & 0.2317 & 0.4858 & \hlsecond{1.9295} & \hlsecond{0.3332} \\
       \midrule
     \multirow{3}{*}{ICT-Net} & 1 & \hlbest{0.4402} & \hlbest{0.3357} & \hlbest{0.2651} & \hlbest{0.2145} & \hlbest{0.2198} & \hlsecond{0.4530} & 1.7041 & 0.2996 \\
     & 3 & \hlsecond{0.4234} & \hlsecond{0.3261} & \hlsecond{0.2605} & \hlsecond{0.2129} & \hlsecond{0.2168} & \hlbest{0.4532} & \hlbest{1.7363} & \hlbest{0.3022} \\
          & 5 & 0.4136 & 0.3192 & 0.2555 & 0.2092 & 0.2144 & 0.4512 & \hlsecond{1.7306} & \hlsecond{0.3008} \\
          \midrule
     \multirow{3}{*}{SparseFocus} & 1 & \hlbest{0.4911} & \hlbest{0.3868} & \hlbest{0.3126} & \hlsecond{0.2570} & \hlbest{0.2458} & 0.5009 & 2.0161 & 0.2515 \\
     & 3 & \hlsecond{0.4725} & \hlsecond{0.3779} & \hlsecond{0.3093} & \hlbest{0.2571} & \hlsecond{0.2438} & \hlbest{0.5060} & \hlbest{2.0761} & \hlbest{0.2579} \\
     & 5 & 0.4652 & 0.3721 & 0.3046 & 0.2536 & 0.2419 & \hlsecond{0.5054} & \hlsecond{2.0624} & \hlsecond{0.2568} \\

    \bottomrule
  \end{tabular}
  }
\vspace{-6pt}
\end{table}
\begin{table}[t]
  \caption{Influence of input image resolution on five representative methods. Higher is better. The \hlbest{best} and \hlsecond{second-best} results in each column are highlighted.}
  \vspace{-6pt}
  \label{tab:res-ablation}
  \centering
  \resizebox{\linewidth}{!}{
  \begin{tabular}{>{\bfseries}cccccccccc}
    \toprule
    \rowcolor{cyan!25}
    \textbf{Method} & \textbf{Resolution} & \textbf{BLEU-1} & \textbf{BLEU-2} & \textbf{BLEU-3} & \textbf{BLEU-4} & \textbf{METEOR} & \textbf{ROUGE\_L} & \textbf{CIDEr} & \textbf{SPICE}\\
    \midrule
    \multirow{3}{*}{RSICCformer} & 256 & 0.3643 & 0.2763 & 0.2175 & 0.1750 & 0.1893 & 0.4083 & 1.3915 & 0.2726 \\
       & 512 & \hlsecond{0.3846} & \hlsecond{0.3000} & \hlsecond{0.2413} & \hlsecond{0.1985} & \hlsecond{0.2028} & \hlsecond{0.4347} & \hlbest{1.6104} & \hlbest{0.2992} \\
       & 768 & \hlbest{0.4070} & \hlbest{0.3163} & \hlbest{0.2525} & \hlbest{0.2054} & \hlbest{0.2101} & \hlbest{0.4452} & \hlsecond{1.6020} & \hlsecond{0.2913} \\
       \midrule
     \multirow{3}{*}{SEN} & 256 & 0.3631 & 0.2689 & 0.2057 & 0.1610 & 0.1827 & 0.3877 & 1.2478 & 0.2669 \\
     & 512 & \hlsecond{0.4385} & \hlsecond{0.3399} & \hlsecond{0.2694} & \hlsecond{0.2163} & \hlsecond{0.2209} & \hlsecond{0.4609} & \hlsecond{1.6644} & \hlsecond{0.3184} \\
          & 768 & \hlbest{0.4690} & \hlbest{0.3717} & \hlbest{0.2997} & \hlbest{0.2456} & \hlbest{0.2405} & \hlbest{0.4984} & \hlbest{1.9164} & \hlbest{0.3501} \\
          \midrule
     \multirow{3}{*}{Chg2Cap} & 256 & 0.4204 & 0.3245 & 0.2596 & 0.2124 & 0.2153 & 0.4495 & 1.6987 & 0.2976 \\
     & 512 & \hlsecond{0.4530} & \hlsecond{0.3585} & \hlsecond{0.2923} & \hlsecond{0.2429} & \hlsecond{0.2347} & \hlsecond{0.4876} & \hlsecond{1.9492} & \hlsecond{0.3345} \\
     & 768 & \hlbest{0.4711} & \hlbest{0.3777} & \hlbest{0.3101} & \hlbest{0.2592} & \hlbest{0.2473} & \hlbest{0.5139} & \hlbest{2.1056} & \hlbest{0.3607} \\
     \midrule
     \multirow{3}{*}{ICT-Net} & 256 & 0.4162 & 0.3208 & 0.2566 & 0.2102 & 0.2147 & 0.4519 & 1.7294 & \hlsecond{0.3042} \\
     & 512 & \hlbest{0.4234} & \hlsecond{0.3261} & \hlsecond{0.2605} & \hlsecond{0.2129} & \hlsecond{0.2168} & \hlbest{0.4532} & \hlsecond{1.7363} & 0.3022 \\
     & 768 & \hlsecond{0.4222} & \hlbest{0.3262} & \hlbest{0.2612} & \hlbest{0.2144} & \hlbest{0.2177} & \hlsecond{0.4530} & \hlbest{1.7444} & \hlbest{0.3060} \\
     \midrule
     \multirow{3}{*}{SparseFocus} & 256 & 0.4284 & 0.3371 & 0.2726 & 0.2241 & 0.2219 & 0.4601 & 1.7816 & 0.2299 \\
     & 512 & \hlsecond{0.4725} & \hlsecond{0.3779} & \hlsecond{0.3093} & \hlsecond{0.2571} & \hlsecond{0.2438} & \hlsecond{0.5060} & \hlsecond{2.0761} & \hlsecond{0.2579} \\
     & 768 & \hlbest{0.5114} & \hlbest{0.4150} & \hlbest{0.3445} & \hlbest{0.2905} & \hlbest{0.2644} & \hlbest{0.5353} & \hlbest{2.3214} & \hlbest{0.2826} \\
    \bottomrule
  \end{tabular}
  }
  \vspace{-14pt}
\end{table}
\subsubsection{Effect of Beam Size}
We further investigate how beam search size during inference affects caption quality. We consider three settings, i.e., 1, 3, and 5, and report results on three representative methods at a fixed resolution of $512 \times 512$. As shown in Table~\ref{tab:beam}, larger beam sizes do not uniformly help. Across methods, beam size 1 generally yields the best BLEU-n and METEOR scores, whereas beam size 3 is usually best on ROUGE\_L, CIDEr, and SPICE, and remains consistently strong overall, which is why we adopt it as the default. Beam size 5 brings no further gain, suggesting that moderate beam search suffices for WUICC.

\rqbox{\textbf{Answer to RQ4.} Performance is sensitive to both factors. Higher input resolution consistently improves results, while a moderate beam size is sufficient and larger beams bring no further gain.}
\section{Discussion}\label{sec:discussion}
\textbf{Implications for VRT practice.} Change captioning supplies the capability VRT pipelines lack. Where a pixel diff returns an undifferentiated mask and floods the pipeline with false positives, a caption directly localizes and names the changed element while suppressing non-meaningful changes more selectively, augmenting VRT triage with a natural language description of what changed.

\textbf{Implications for future modeling.} Web UI changes are small relative to the canvas and frequently textual, so progress depends on capabilities the natural image and remote sensing benchmarks do not exercise, namely verbatim text reading and fine-grained localization of a small changed region within a large unchanged canvas~\cite{li2025region}. Building the WUICC-specific architecture these axes call for is left to future work that our benchmark is designed to enable.

\subsection{Threats to Validity}\label{sec:limitations}
\textbf{Construct validity.} Our rankings rely on captioning metrics (BLEU, METEOR, ROUGE\_L, CIDEr) that reward surface lexical similarity. Verbatim reproduction of text edits is partially captured, since CIDEr and high-order BLEU reward exact, TF-IDF-weighted matching of the changed text and penalize substituted tokens, whereas METEOR and SPICE credit synonyms and so do not reflect verbatim correctness. Even so, because the changed text is a small fraction of a caption, a single-token error is diluted. These metrics nonetheless remain the standard, reproducible basis for comparison in IDC, and our conclusions do not rest on any single one.

\textbf{Internal validity.} Because the ground-truth captions are produced by the LLM-driven pipeline rather than written by humans, models scored under lexical metrics could be favored for sharing the generator's phrasing. We mitigate the circularity in three ways. First, trained annotators verify every sample, checking that the caption completely and accurately describes the change in the rendered pair, including verbatim correctness of any changed text, with substantial agreement ($\kappa \approx 0.722$) and 78.09\% of generations accepted, so fitting these references means learning accurate change descriptions. Second, the bias cannot affect the rankings our conclusions rest on, since all eleven benchmarked methods are trained from scratch on the same references and the two zero-shot baselines (Llama and Qwen) come from a different family than the GPT-family generator, which leaves only the absolute scores at risk. Third, our change and no-change accuracies reward correct content rather than wording, and corroborate the metric-based trends.

\textbf{External validity.} Our source pages come from WebSight~\cite{laurenccon2024unlocking}, a corpus of LLM-synthesized HTML curated for general UI training rather than the framework-heavy production sites where visual regressions matter most, so the layouts, components, and dynamic JavaScript-driven content of WUICC-bench need not match real production UIs. Each sample also contains exactly one taxonomy-drawn change, a deliberate choice for clean attribution and tractable verification rather than an oversight (Section~\ref{sec:taxonomy-rationale}), whereas real regressions sometimes involve multiple interacting changes within a single release, an extension we leave to future work. Finally, our coverage of LLM-based change captioning is partial. Specialized systems such as Change-Agent~\cite{liu2024change}, ChangeChat~\cite{deng2025changechat}, and TEOChat~\cite{irvinteochat} require supervision from change masks or bounding boxes that WUICC-bench does not provide, so we do not evaluate them. We do include two open general-purpose VLMs, Llama-3.2-11B-Vision-Instruct and Qwen2-VL-7B-Instruct, as zero-shot baselines, and find that they fall short of the trained methods on both caption quality and selective suppression. A broader sweep of stronger and closed-source VLMs such as GPT-5 and Claude is costly to query with two high-resolution images, and their zero-shot quality varies widely across models, so a systematic study of these larger VLMs is left to future work. Quantifying the production gap with a held-out probe set and extending the pipeline to compound changes are also next steps.
\section{Related Work}\label{sec:related}
\textbf{Visual Regression Testing.}
Academic VRT research reduces the manual confirmation burden of pixel-level comparison through grouping identical differences~\cite{adachi2018reducing}, region-based matching~\cite{tanno2020region}, masking dynamic content areas~\cite{adachi2020method}, and hybrid image-HTML comparison~\cite{aridome2026mixvrt}, with a complementary line on dependent tests that reduces flaky failures~\cite{lam2020dependent}. These output binary flags or highlighted regions, whereas our work produces a natural language description that such detection-oriented techniques cannot. Recent VLM work detects cross-browser inconsistencies~\cite{xbidetective2026} and repairs visual issues~\cite{huang2025seeing}, but targets detection and repair rather than describing the change.

\textbf{Remote Sensing Image Change Captioning.}
RSICC research has progressed through several methodological waves: early pipelines that couple a convolutional visual encoder with a recurrent or support vector machine based caption generator~\cite{chouaf2021captioning,hoxha2022change,sun2025scene,li2025cd4c,zou2025frequency,peng2024change,yang2025restricted}, diffusion-based generation~\cite{yang2024remote,yu2025diffusion,sun2025mask,bai2025cross}, Transformer- and attention-based architectures~\cite{sun2024lightweight,wu2025cross,chang2023changes,liu2022remote,shi2024multi,zhou2024single,cai2023interactive,chen2024multi,karaca2025robust,hang2025text}, Mamba state-space models for linear-complexity sequence modeling~\cite{ma2025cross,liu2024rscama,ma2025ihm}, and most recently LLM-based approaches~\cite{liu2023decoupling,liu2024change,irvinteochat,bazi2024rs,yang2025enhancing,deng2025changechat,zhu2024semantic}. Our benchmark adopts representative methods from each wave as baselines.

\textbf{Image Difference Captioning.}
IDC on natural images has explored change-region identification~\cite{qiu2021describing}, contrastive pre-training~\cite{yao2022image}, cross-view representation reconstruction~\cite{tu2023self}, viewpoint-robust disentanglement~\cite{tu2023adaptive}, common-difference feature decoupling~\cite{tu2024context}, regional difference modeling~\cite{li2025region}, multi-scale perception in multimodal LLMs~\cite{liu2025omnidiff}, prompt-to-prompt synthetic pairs~\cite{black2024vixen}, generalist VLMs with visual-delta modules~\cite{hu2024onediff}, BLIP2 adaptation~\cite{evennou2025reframing}, and describing differences across image sets~\cite{dunlap2024describing}. Likewise, we select representative methods for our benchmark.
\section{Conclusion}\label{sec:conclusion}
WUICC-bench is the first dataset and benchmark for WUICC, the task of describing Web UI changes in natural language for VRT, built by a controlled LLM-driven mutation pipeline with human-verified captions. Evaluating eleven representative methods, we find that existing models transfer poorly to the Web UI domain, yet already suppress non-meaningful changes more selectively than pixel-level comparison, the capability VRT most needs. Our results position change captioning as a promising paradigm for semantic, more informative VRT and a foundation for future domain-specific methods.




\bibliographystyle{IEEEtran}
\bibliography{IEEEexample}

\end{document}